\documentclass{llncs}
\usepackage{amsmath}
\usepackage{graphicx}
\usepackage{xcolor}
\usepackage{array}
\usepackage{enumerate}

\usepackage{algorithm}
\usepackage[noend]{algpseudocode}

\algnewcommand\algorithmicforeach{\textbf{for each}}
\algdef{S}[FOR]{ForEach}[1]{\algorithmicforeach\ #1\ \algorithmicdo}
\newcolumntype{P}[1]{>{\centering\arraybackslash}p{#1}}
\begin{document}
\title{Mining Human Mobility Data to Discover Locations and Habits}
%
%\titlerunning{ECML/PKDD 2019 Workshop on IoT Stream for Data Driven Predictive Maintenance}
\author{Thiago Andrade\inst{1,2}, {Brais Cancela}\inst{1,3}
\and Jo\~ao Gama\inst{1,2}
}
%\authorrunning{Andrade et al.}
\institute{INESC TEC, Porto, Portugal \and
University of Porto, Porto, Portugal \and
Universidade da Coru\~na, Spain\\
\email{thiago.a.silva@inesctec.pt}
}
\maketitle              % typeset the header of the contribution
\begin{abstract}
Many aspects of life are associated with places of human mobility patterns and nowadays we are facing an increase in the pervasiveness of mobile devices these individuals carry. Positioning technologies that serve these devices such as the cellular antenna (GSM networks), global navigation satellite systems (GPS), and more recently the WiFi positioning system (WPS) provide large amounts of spatio-temporal data in a continuous way. Therefore, detecting significant places and the frequency of movements between them is fundamental to understand human behavior. In this paper, we propose a method for discovering user habits without any a priori or external knowledge by introducing a density-based clustering for spatio-temporal data to identify meaningful places and by applying a Gaussian Mixture Model (GMM) over the set of meaningful places to identify the representations of individual habits. To evaluate the proposed method we use two real-world datasets. One dataset contains high-density GPS data and the other one contains GSM mobile phone data in a coarse representation. The results show that the proposed method is suitable for this task as many unique habits were identified. This can be used for understanding users' behavior and to draw their characterizing profiles having a panorama of the mobility patterns from the data.
\end{abstract}
\section{Introduction}
\label{sec:Introduction}

Understanding human mobility patterns can help in the exploration of the underlying driving factors of society as many aspects of life are associated with them. The first efforts to learn human mobility patterns were associated with classic social sciences. Since the nineteenth century, sociologists in what are called time-use or time-budget studies have been measuring the time people spend doing different activities throughout the day \cite{toch2019analyzing}. In contrast, methods for human mobility data collection have shifted over time as now both developed and developing countries are facing the increase of the pervasiveness of mobile devices \cite{berry2011computational,lazer2009computational}. Positioning technologies that serve these devices such as the cellular antenna (GSM networks), global navigation satellite systems (GPS), and more recently the WiFi positioning system (WPS) provide large amounts of spatio-temporal data in a continuous way at low costs \cite{liu2007survey}.
When dealing with raw data, final users cannot make sense of it without processing and applying techniques to extract meaningful information from its content. Many researchers have made efforts in exploring these data in order to find places, locations, and regions \cite{lee2013mining,zheng2010geolife,zheng2009mining}. Hence, individuals can state a place as something with a meaning such as work, home, university while a pair of numbers like "39.98450, 116.29929" has no useful meaning to them. Therefore, detecting significant places and the frequency of movements between them is fundamental to understand human behavior.

%In this way, as Calabrese surveyed in \cite{calabrese2015urban}, different aspects of cellular network data have been used for different researches. González \cite{gonzalez2008understanding} and Song \cite{song2010limits} analyzed mobility patterns and urban mobility is discussed in \cite{calabrese2011real,alhasoun2014city}.

Several studies confirmed the intuition that human mobility is highly predictable, centered on a small number of base locations \cite{herder2012daily}. This opens a wide range of opportunities for more intelligent recommendations and support for routine activities. Still, empirical studies on individual mobility patterns are scarce.

The main contributions of this paper are related as follows: we introduce a new dataset acquired from a Telecom company that comprises many different cities in Brazil. We also present a new density-based clustering for spatio-temporal data to identify meaningful places. Moreover, in the last step, we apply a Gaussian Mixture Model (GMM) over the Origin x Destination matrix of trips between meaningful places to automatically separate the trajectories for identification of user habits.

The following section presents the literature review and the most important related works. The remainder of the paper describes the methodology and the data sets utilized to assert the validity of the methods in section \ref{sec:Methodology}, in section \ref{sec:ExperimentsResults} we discuss the experiments and results obtained. Finally, the conclusions and future work are presented in section \ref{sec:ConclusionFutureWork}.
\section{Related Work} 
\label{sec:RelatedWork}

Many researchers have been proposing methods to identify meaningful locations and habits from users for diverse goals. In this section, we review some relevant works which leverage the information contained in GPS and mobile phone data (GSM) for a multitude of different applications.

According to \cite{lee2013mining}, several methods based on density have been proposed in order to discover regions of interest although most of these methods are used to aggregate spatial point objects.
Some authors were more interested in the semantic movement trajectories. \cite{li2008mining} introduced a model that makes use of movement datasets which has trajectories defined as sequences of time-stamped stops and moves between locations. 
In order to discover personalized visited-POIs, \cite{suzuki2019personalized} proposed a method to estimate fine-grained and pre-defined locations. 
In \cite{andrade2018pois} the authors explore raw GPS data to identify meaningful places in a region and describe user's profiles and similarities among them. %The proposed method on density-based clustering to identify points of interest (POIs) and uses Jaccard similarity to list the users that are alike.

Many researchers were also interested in mobility patterns.
Most location-based services provide recommendations based on a user's current location or a given route or destination. Even though there are indications that human movement is highly predictable, daily and weekly routines of individual users constitute a largely unexplored and unexploited area.
\cite{thuillier2018clustering} used more than 800 million of CDR data to identify weekly patterns of human mobility through mobile phone data. %They proposed a methodology based on the classification of individuals into presence profiles focusing on the inherent temporal and geographical characteristics.
In \cite{sardianos2018extracting}, the authors present a methodology based on density-based clustering, clustering-based sequential mining and Apriori algorithm for analyzing user location information in order to identify user habits.
%
%
%%%%%%%%%%%%%Methodology%%%%%%%%%%
\section{Problem Statement and Methodology}
\label{sec:Methodology}
The objective of this work is to propose a methodology to identify user habits from GPS and GSM data without any apriori or external information. We propose a variation of DBSCAN clustering technique that is able to perform cluster of locations like buildings and squares in a better way and apply a GMM in order to separate the days and hours a given user moves between the clustered locations.

Before entering in details of the methodology, we introduce the definition of points and trajectories:\\
A point is a triple of the form \textit{p} \texttt{=} \textit{(latitude, longitude, time)} that represents a latitude-longitude location and a time-stamp.
A trajectory is a sequence of ordered points triples \textit{Tr} \texttt{=} $(p_1$, $p_2$, . . . , $p_n)$ where $p_i$ is a point and $p_1.time < p_2.time < . . . < p_n.time.$

The first step of the methodology is the preprocessing task that is including among other activities, the data cleaning process where we perform outliers and noise removal. The second is the feature engineering to derive new information from the original data (in the form of latitude, longitude, and a time-stamp) to calculate key features such as time delta of the transitions, traveled distance between points, velocity, start and stop positions, time and day of the week, length and duration of a trajectory. In this work, we denote a new trajectory every time an individual stop moving or the time delta between points is  more than 30 minutes.

%%%%%%%%%%%%%UserStayPointsDetection%%%%%%%%%%
\subsection{User Stay Points Detection}
\label{subsec:UserStayPointsDetection}

Stay points are regions where a given user has stayed for a while within a defined radius. 
The algorithm is a hybrid density and time-based proposed in \cite{ye2009mining} that calculates the distance between two sets of points $p_1$ and $p_0$ in order to find those that are below a distance threshold. Next, it checks for how long the user stayed in that radius by looking at time threshold. At last step, it calculates the stay points centroid by getting the mean of the coordinates of the set of points. For this experiment, we set the parameters Distance-threshold as 200 meters and the Time-threshold to 20 minutes as suggested in \cite{zheng2009mining}. %We can see an example of stay point in the Figure \ref{fig:stayPointsDef}

%\begin{figure}[!h]
%\centering
%\includegraphics[width=0.5\textwidth]{images/stayPoints}
%\caption{GPS log and stay points \cite{ye2009mining}} \label{fig:stayPointsDef}
%\end{figure}

%%%%%%%%%%%%%UserStayPointsDetection%%%%%%%%%%

%%%%%%%%%%%%%MeaninfgulLocations%%%%%%%%%%
\subsection{Meaningful Locations}
\label{subsec:MeaninfgulLocations}

A meaningful place is defined as a frequent location visited by an individual and does not need to be related to any other person or group like in the case of the POIs. Taking into account we already have the user's stay points, now we need to look for those places (stay points) a person visits repeatedly in order to form the so-called users' meaningful places.

Location detection techniques commonly make use of density-based methods. This is because the mechanism of density-based clustering is able to detect clusters of arbitrary shapes without specifying the number of the clusters in the data a priori and is also tolerant of outliers (noise).

The Location Clustering method proposed by \cite{ashbrook2003using}, operates attributing in a way that once it forms a cluster, these points are eliminated from the neighbourhood and avoid new points to overlap to them. In this way, the remaining observations are available to form new clusters surrounding the previous center that could maybe be part of it. Our method, on the other hand, keeps a short memory for those points revisiting and maybe reclassifying them to the new cluster as the density of the new class turns to be more relevant.

One main advantage over the classical DBSCAN \cite{ester1996density} implementation is that given the arbitrary shape of the trajectories, sometimes the clusters form straight chains which may not be a good representation of a location as normally buildings are in a squared or circular shape. Our method is robust to these situations as it classifies as noise those points that fall out of the neighbour's radius which is away from the centroid of the cluster.
Another drawback of this original DBSCAN approach is that it does not return a centroid for each cluster. As we are looking for meaningful places over the set of stay points (section \ref{subsec:UserStayPointsDetection}), we need to find the centroid for each of the returned labels of the DBSCAN. 

To overcome these issues, we propose a variation of the clustering algorithm DBSCAN \cite{ester1996density} and Location Clustering \cite{ashbrook2003using} methods. The method starts searching for a given $p$ point neighbours ($MinPts$) in as $Eps$ radius. While the set of neighbours still changing, it keeps on looping through the data points. Once it stops changing, it checks if the number of items in the class is greater than the minimum points to form a cluster. If this condition holds, we set all the points into this given neighbourhood to noise and move the centroid of the list of points to iterate over again.
The algorithm proposed is described the pseudo-code~\ref{alg:DBSCANKmeans}.

\begin{algorithm}[!hbt]
\caption{DBMeans Algorithm}\label{alg:DBSCANKmeans}
%\algsetup{linenosize=\tiny}
\scriptsize
\begin{algorithmic}[1]
\Function{DBMeans}{$P, eps, MinPts$}

\State \% $P$: a set of points (lat, long)
\State \% $eps$: the radius of the cluster $>$ 0
\State \% $MinPts$: the minimum size of a cluster $>$ 0
\State \% $C$: the label of each point in $P$
\State \text{$C \leftarrow$ NOT VISITED}\;
\State \text{$Centroids \leftarrow []$}\;

\While{$\exists \hspace{0.1cm} p_c \in P | C(p_c) = \text{NOT VISITED}$}:

    \State \text{$M \leftarrow \{p_c\}$}\;
    \State \text{$M_{bak} \leftarrow \{\emptyset\}$}\;
    \State \text{$C(p_c) \leftarrow $ NOISE}\; \Comment{Mark $p_c$ as noise}
    \State \text{$c$ \texttt{=} $p_c$}\; \Comment{Random choice from shuffled input P}
    
    \While{$M \neq M_{bak}$}
        \State \text{$M_{bak} \leftarrow M$}\;
        \State \text{$M \leftarrow \{\emptyset\}$}\;
        \State \text{$C_M \leftarrow []$}\;
        \ForEach{$p_x \in P$}:
            \If{$distance(c, p_x) < eps$}
            	\State $M \leftarrow M \cup \{p_x\}$\;
            	\If{$C(p_x) \notin \{ NOT VISITED, NOISE \}$} 
            	    \State $C_M \leftarrow C_M \cup \{C(p_x)\}$\;
            	\EndIf 
            \EndIf 
        \EndFor
        \State $c \leftarrow Mean(M)$\;
        \ForEach{$c_m \in set(C_M)$}:
            \If{$| c_m | \geq MinPts$} 
            	%\State $M \leftarrow M \cup \{p_x\}$\;
            	\State $C_M(C_M = c_m) \leftarrow$ NOISE\; \Comment{Mark $c_m$ as noise}
            	\State $Centroids \leftarrow Centroids \setminus c_m$\; 
            \EndIf 
        \EndFor
    \EndWhile
    \State $C(M) \leftarrow c_c$\; \Comment{Mark all neighbour points with the same class}
    \State $Centroids(c_c) \leftarrow c$\;
\EndWhile
\State \text{$C \leftarrow predict(P, Centroids, eps)$}\;
\State \textbf{return} $C$
\EndFunction
\end{algorithmic}
\end{algorithm}
%\clearpage

%%%%%%%%%%%%%MeaninfgulLocations%%%%%%%%%%

%%%%%%%%%%Identification of Habits%%%%%%%%%%%%%%%
\subsection{Identification of Habits}
\label{sec:IdentificationOfHabits}

Individuals have a remarkable propensity to return to their frequently visited places. Hence, the interactions between individuals and these places are likely to represent the individual's characteristics. %As mentioned in the section \ref{subsec:Preprocessing}, some features were derived from the original datasets in order to perform the knowledge discovery process.
After clustering the user stay points into meaningful places as described in \ref{subsec:MeaninfgulLocations} we ended up with: trajectories connecting non-meaningful places (those who start and end in places classified as noise), trajectories connecting one meaningful place at the end or at the start and trajectories connecting two meaningful places. For the habits study purpose, we will focus on the last item as we are interested in discovering frequent movements across meaningful places.

From this list of grouped trajectories is possible to identify the most important places of a given user as we can perform a count on the occurrences of trips connecting two locations. Groups with very low values, close to zero, means that there are no habits connecting those places or the $eps$ parameter used to perform the clustering in step \ref{subsec:MeaninfgulLocations} is too small. For this study, we are considering only the two locations that have at least 5 (five) trajectories connecting them.

\subsection{Gaussian Mixture Model to classify the different habits}
\label{sec:GMM}

In order to discover user habits, we need to analyze the features that are emerging from the discovery process. One way we can utilize to separate the trips into habits is by the time they happen.
%Some data is inherently cyclical. Time is a rich example of this: minutes, hours, seconds, day of week, week of month, month, season. The time is continuous and circular whereas the day of the week is a discrete ordered value and with daily user habits its not different. Indeed, 23:00pm is as far as 1:00am than 21:00pm. Our idea is to transform the start and end hour of each trajectory in a way that 23:55 and 00:05 are 10 minutes apart and may be part of a same habit. 
To tackle this issue we create two new features, deriving a sine and cosine transform from the start hour.

\begin{figure}[!h]
\centering
    \includegraphics[width=0.40\textwidth]{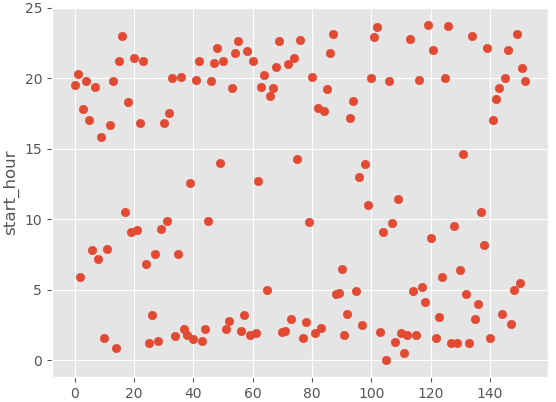}
    \includegraphics[width=0.3\textwidth]{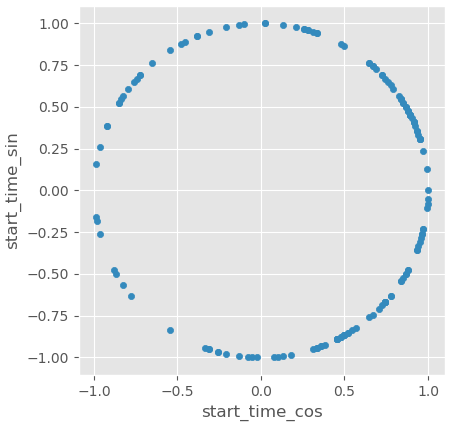}
\caption{Transformation of the start hour based on the Sin and Cos. The left image represents the hours in a plain representation (X axis is the trajectory order), the right is a circular where two or more points can fall over the same region no mater the trajectory order} \label{fig:clock}
\end{figure}

In figure \ref{fig:clock} we show an example of the transformation based on the start hours all trajectories of a user to show the new representation of time. 

\begin{figure}[!h]
\centering
    \begin{minipage}[]{0.28\linewidth}
        \centering
        \includegraphics[width=0.99\textwidth]{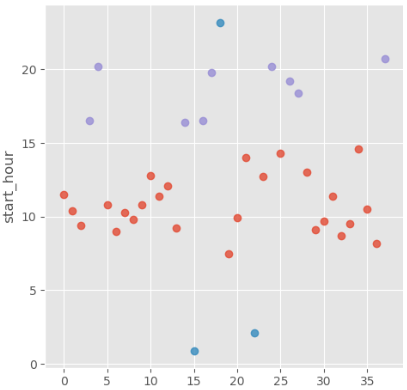} \\
        (a) Starting hours and their classes.
    \end{minipage}
    \begin{minipage}[]{0.34\linewidth}
        \centering
        \includegraphics[width=0.99\textwidth]{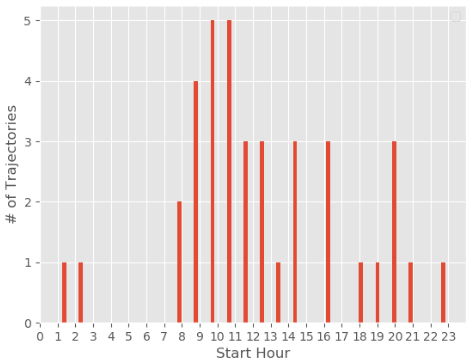} \\
        (b) Histogram of the starting hours with three main peaks.
    \end{minipage}
    \begin{minipage}[]{0.36\linewidth}
        \centering
        \includegraphics[width=0.99\textwidth]{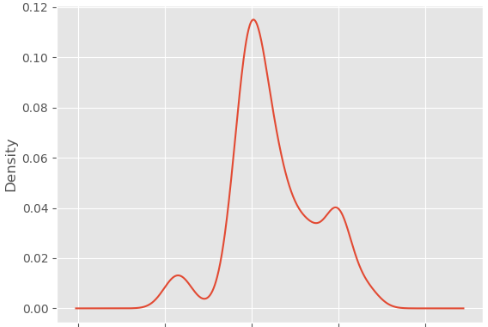} \\
        (c) Densities of the mix of Gaussians found over the distribution.
    \end{minipage}
\caption{GMM model over the start hours of the trajectories.} \label{fig:gmm}
\end{figure}

The cyclical representation of the time is not enough when dealing with individuals that use to go to a certain place in a non-strict way. The distribution of the data may be non-normal resulting in more than one peak along the day. Here we propose utilizing a Gaussian Mixture Model to handle these cyclical data and segment it into habits in a dynamic way. 
Figure \ref{fig:gmm} shows the starting hours of a given user habit. One can notice that this user has 37 different starting hours for the same Origin x Destination pair and is possible to verify the segmentation made by the multiple Gaussians in the start hour distribution. Note that there are blue dots on the top and the bottom of the left image representing the same class of trajectories that occur close to 23:00pm and 02:00am.

%%%%%%%%%%Identification of Habits%%%%%%%%%%%%%%%
%%%%%%%%%%%%%Methodology%%%%%%%%%%

%%%%%%%%%%Experiments and Results%%%%%%%%%%%%%%%
%
%
\section{Experiments and Results}
\label{sec:ExperimentsResults}

In this section, several experiments with the two real-world datasets are performed to evaluate our proposed method. The datasets description and their preparation are described in sub-subsections \ref{subsubsec:Geolife}. Subsection \ref{subsec:ClusteringResults} corresponds to clustering results and subsection \ref{subsec:HabitsResults} presents the results regarding the habits extraction.

\subsection{Datasets}
\label{subsec:Datasets}
\subsubsection{Geolife GPS Dataset}
\label{subsubsec:Geolife}
%%%%%%%%%%%%%Dataset%%%%%%%%%%
This GPS trajectory dataset was collected in (Microsoft Research Asia) Geolife project by 182 users in a period of over three years (from April 2007 to August 2012). 
The dataset contains 17,621 trajectories with a total distance of about 1,2 million kilometers and a total duration of 48,000+ hours. These trajectories were recorded by different GPS loggers and GPS-phones, and have a variety of sampling rates. 90\% of the trajectories are logged in a dense representation, e.g. every 1 to 5 seconds or every 5 to 10 meters per point \cite{zheng2008understanding,zheng2009mining,zheng2010geolife}.

\subsubsection{GSM Telecom dataset}
\label{subsubsec:GSMTelecom}
%%%%%%%%%%%%%Dataset%%%%%%%%%%
This is a new dataset based on mobile phone (GSM) data. The dataset contains 526,894 instances from a period of 12 months or 350 days starting on September 2017 and finishing in September 2018 consisting of 4,545 different individuals. After cleaning and removing the duplicates, it was reduced to 461,778 instances. The points were recorded in many cities in Brazil with a coarse granularity of one point at every 15 minutes. No information about the users is derived from these data, as the entire dataset is anonymized. 
Each point consists of a user sequential identification number, a pair of (latitude, longitude), and a timestamp. All the data was delivered in a single file that is available in the project folder on the web page.\protect\footnote{ \protect\url{https://bit.ly/2ZVERKO}}
%%%%%%%%%%%%%Dataset%%%%%%%%%%

\subsection{Clustering Results}
\label{subsec:ClusteringResults}

\begin{figure}[!h]
\centering
    \begin{minipage}[]{0.31\linewidth}
        \centering
        \includegraphics[width=0.99\textwidth]{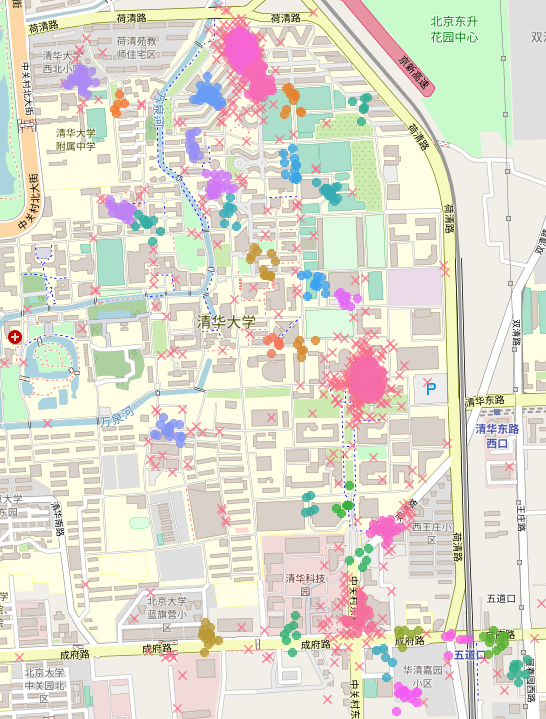} \\
        (a) Our method returns more concise clusters as it doesn't erase the nearby points after forming a cluster. The surrounding points are set to noise when the mean of the points inside the radius stops changing.
    \end{minipage}
    \begin{minipage}[]{0.32\linewidth}
        \centering
    \includegraphics[width=0.99\textwidth]{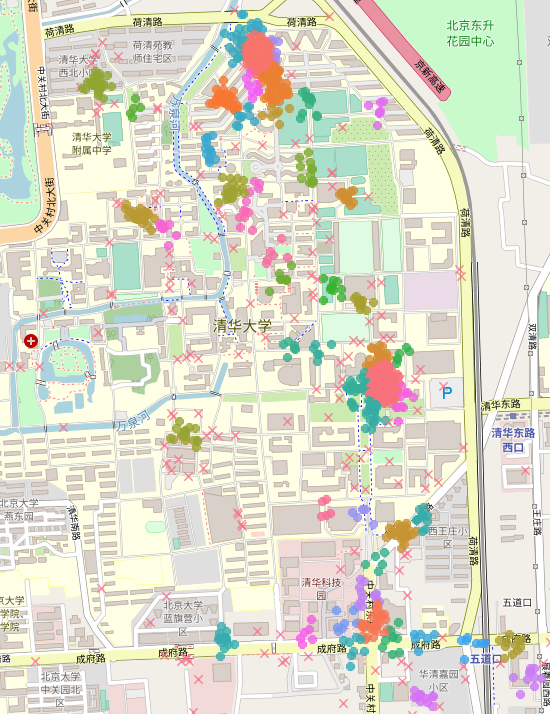} \\
        (b) Location Clustering: returns satellite clusters over the  main location as it works erasing those points who form a cluster after the mean stops changing. Noisy points can be set to a possible new cluster as can be seen in detail.
    \end{minipage}
    \begin{minipage}[]{0.32\linewidth}
        \centering
        \includegraphics[width=0.99\textwidth]{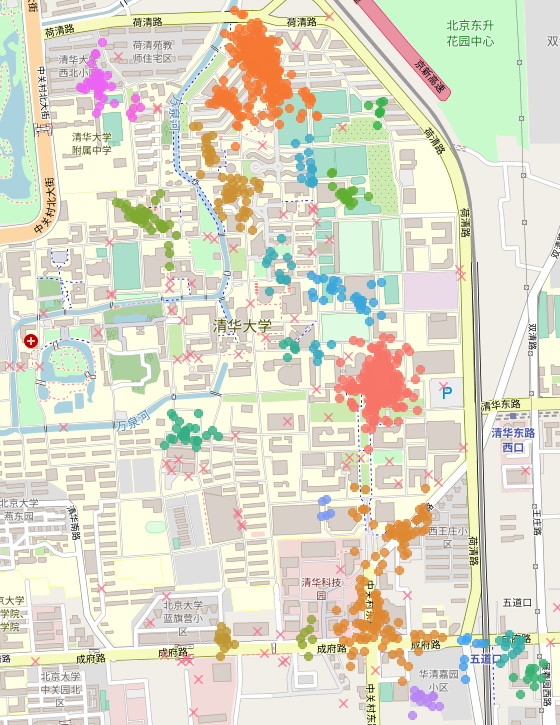} \\
        (c) DBSCAN: forms very large clusters from chaining points which are density-connected. This is of the main disadvantage in this context as the shape of the locations is usually in squared or circular different from ellipses. 
    \end{minipage}
\caption{The dense region in the top shows the clear difference among the methods: while our approach (a) returns only two clusters, the Location Clustering (b) returns 9 and DBSCAN (c) returns only one large cluster. The $X$ symbol stands for noise} \label{fig:clustersGeolife}
\end{figure}

Following we present the results of the experiments over the two datasets with respect to the identification of Meaningful Places.
%bianchi2016identifying,
To conduct the experiments over the Geolife dataset we elected the individual '004' who seems to be an average person. This user has 1.100 trajectories starting from 2008-10-23 to 2009-07-28 in which are related to 2.437 stay points. From those stay points, 50 meaningful places (MPs) were identified by using the clustering method proposed in \ref{subsec:MeaninfgulLocations}. The top two MPs are latitude\textit{=}$39.99993$, longitude\textit{=}$116.32730$ which has 659 visits, and latitude\textit{=}$40.01086$, longitude\textit{=}$116.32186$ with a counting of 235 times. Here we set the Home (Qinghuayuan Residential District) and Work (Tsinghua University Northwest) locations respectively based on the frequency of these observations as many other works propose \cite{ashbrook2003using,herder2012daily,sardianos2018extracting,yang2018mining}. To perform a visual inspection of the formed clusters, figure \ref{fig:clustersGeolife} illustrates the differences obtained using each one of the methods. Notice that our approach results in clusters that are more robust and handle the noise with more efficiency.

\begin{figure}[!h]
\centering
    \includegraphics[width=0.69\textwidth]{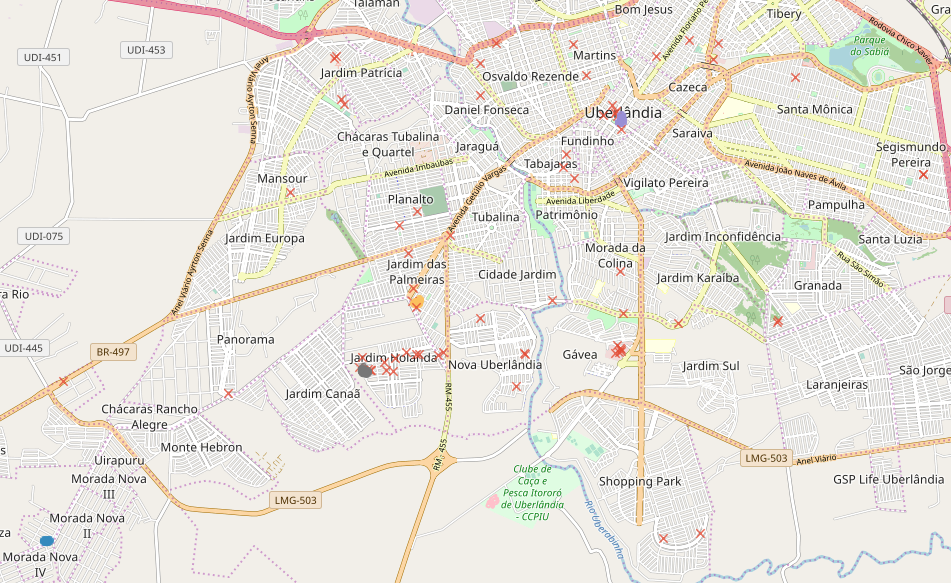}
\caption{Meaningful locations (colored circles) of the user 10837 over the Uberl\^andia/Brazil map. The $x$ symbol stands for noise} \label{fig:clustersGSM}
\end{figure}

Regarding the GSM Mobile Telecom dataset, as in this dataset the granularity is coarse, the results are quite different from the ones shown in the GPS dataset as we have one observation at every 15 minutes. Although Brazil is a very large and populated country, the latitudes and longitudes encountered in this dataset fall into some very small up to medium cities with traffic conditions very different from Beijing. One can notice that in this case, a 15-minute interval can lead to the transportation of the individual to a very different location without any details of the trajectory taken. Basically, we end up with the start and end of the trajectory only.
To conduct the experiments over this dataset we elected the individual '10837'. This user has 19 trajectories starting from 2018-05-10 to 2018-07-01 in which are related to 135 stay points from 4 meaningful places (MPs). The top two MPs are latitude\textit{=}$-18.96081$, longitude\textit{=}$-48.32141$ which has 38 visits, and latitude\textit{=}$-18.94969$, longitude\textit{=}$-48.31219$ with a counting of 6 times.
In figure \ref{fig:clustersGSM} we can see the locations over the map of Uberl\^andia/Brazil.

\subsection{Habits Results}
\label{subsec:HabitsResults}

\begin{figure}[!h]
\centering
\includegraphics[width=0.75\textwidth]{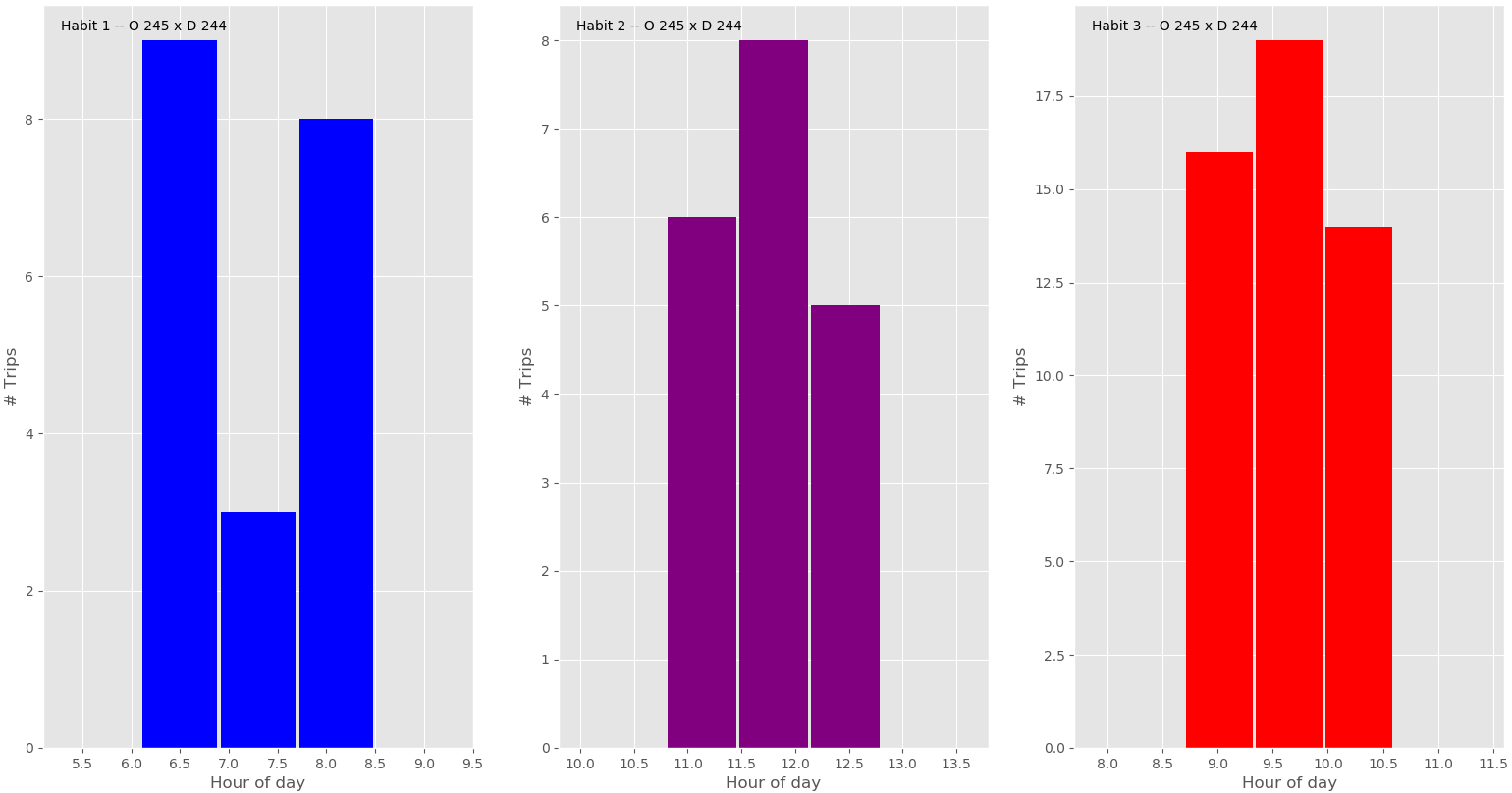}
\caption{Three main habits returned from the start hours connecting the top two locations of the user '004'} \label{fig:habits_compared}
\end{figure}

\begin{figure}[!h]
\centering
\includegraphics[width=0.36\textwidth]{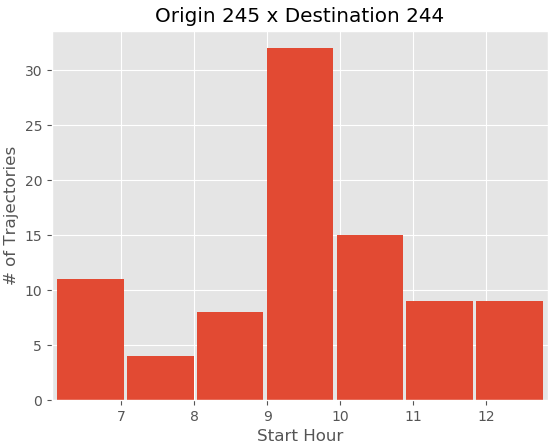}
\includegraphics[width=0.36\textwidth]{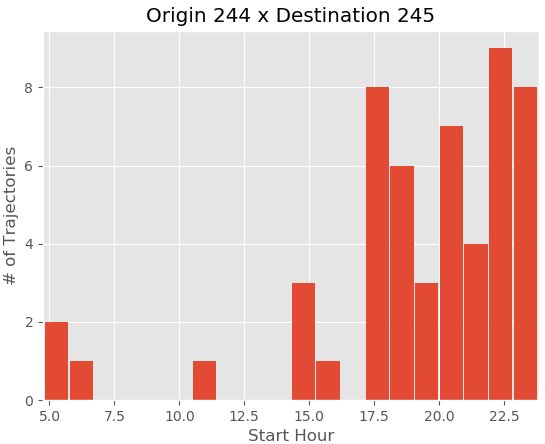}
\caption{Distribution of starting hours from the top two meaningful locations of the user '004'} \label{fig:od_pairs_hour}
\end{figure}

The knowledge discovery process over raw location data has led us to a panorama of the mobility patterns of the given community. Main factors that characterize habits are related to the start hour, length and duration of the trajectories that follow an Origin x Destination pattern. The graph \ref{fig:habits_compared} shows the three different habits returned from the Gaussians for the trajectories of the user connecting the two locations. The graph \ref{fig:od_pairs_hour} illustrates the hourly distribution of the trajectories between the two main groups of meaningful places for the Geolife dataset user '004'.

\begin{figure}[!h]
\centering
\includegraphics[width=0.36\textwidth]{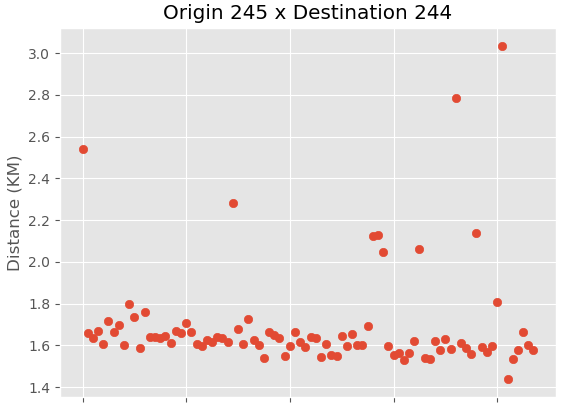}
\includegraphics[width=0.36\textwidth]{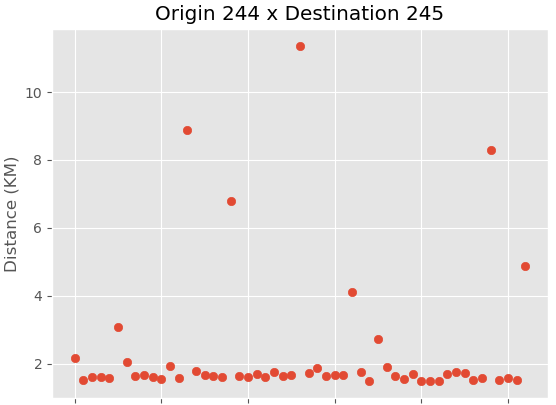}
\caption{Distribution of trajectory distance from the top two meaningful locations of the user '004'} \label{fig:od_pairs_length}
\end{figure}

The length of the trajectories is also a discriminant feature, as users tend to follow the same path to go from places according to evaluated conditions such as day of the week, the hour of the day, weather. In rush hours is more reasonable to avoid areas with too many people and traffic as the time taken to run the same path can be completely distinct. The graph \ref{fig:od_pairs_length} shows the length of the trajectories with respect to their groups of meaningful places. As we can see, some few trajectories have distance greater than 2km. Those can be justified as non-habitual paths.

\begin{figure}[!h]
\centering
\includegraphics[width=0.36\textwidth]{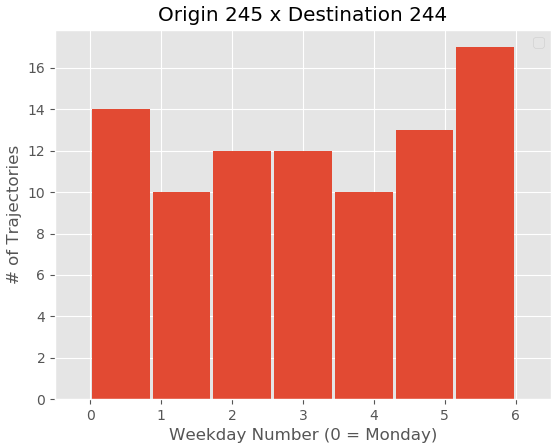}
\includegraphics[width=0.36\textwidth]{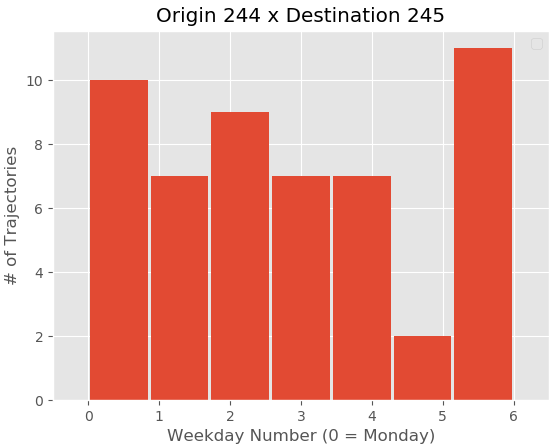}
\caption{Distribution of the trajectories according to the day of week from the top two meaningful locations of the user '004'} \label{fig:od_pairs_wk}
\end{figure}

\begin{figure}[!h]
\centering
\includegraphics[width=0.24\textwidth]{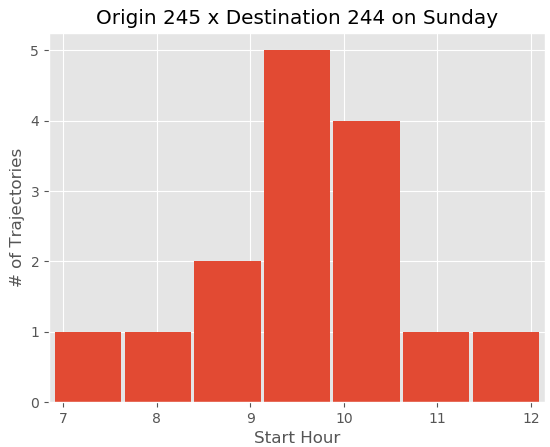}
\includegraphics[width=0.24\textwidth]{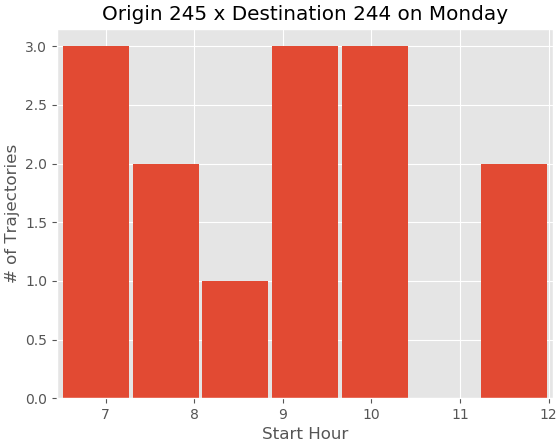}
\includegraphics[width=0.24\textwidth]{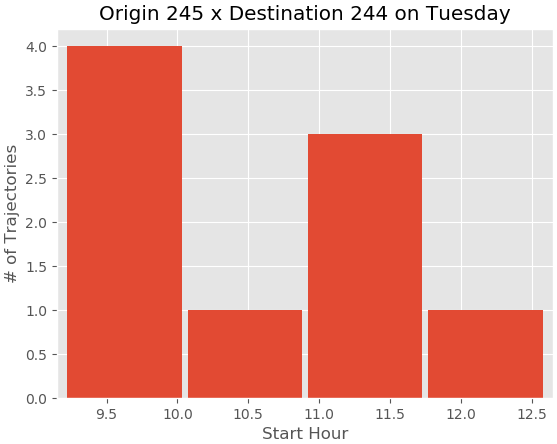}
\includegraphics[width=0.24\textwidth]{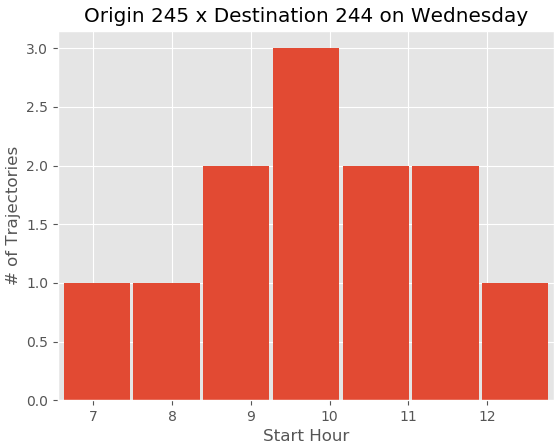}
\includegraphics[width=0.24\textwidth]{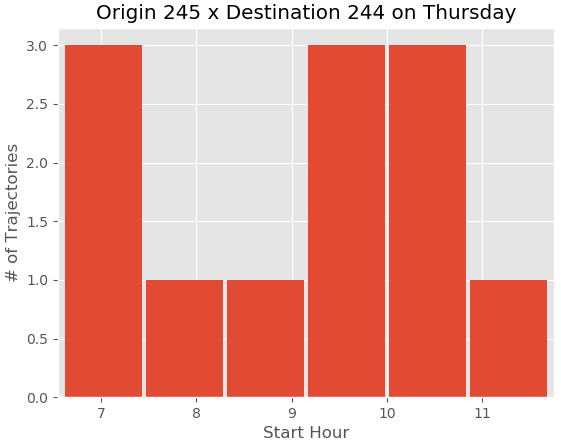}
\includegraphics[width=0.24\textwidth]{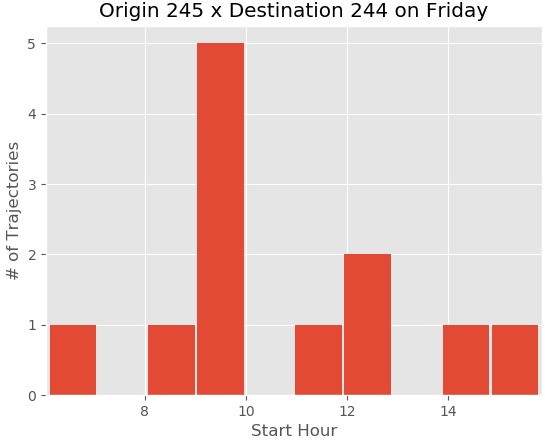}
\includegraphics[width=0.24\textwidth]{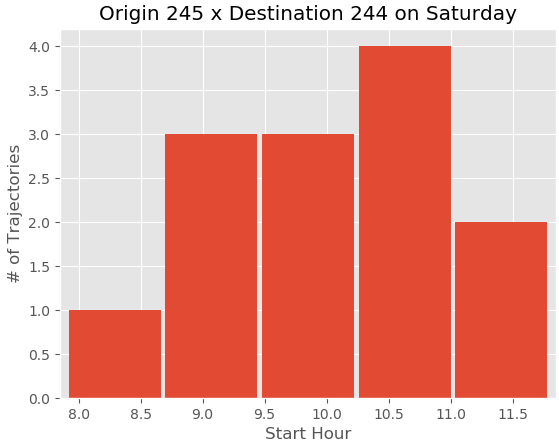}
\caption{Distribution of the trajectories according to the hour and day of week from the top two meaningful locations of the user '004'} \label{fig:od_pairs_wk_d}
\end{figure}

Another  relevant way to analyze habits is looking for the day of the week a trip was taken. Routines are very common in human patterns and some of them may occur less often than the others. In graphs \ref{fig:od_pairs_wk} and \ref{fig:od_pairs_wk_d} we show the distribution of the trips in a weekly view.

%To have an idea of the spatial distribution of the top 5 locations of this given user we zoomed the area they fall into the Beijing map as presented in Figure \ref{fig:top_locations}. All those locations are related to the Tsinghua University which composes a very large area in the northwestern region of the city.

%\begin{figure}[!h]
%\centering
%    \includegraphics[width=0.75\textwidth]{images/004_top5}
%\caption{Panorama of the user's '004' habits where the top 5 locations are showed. The more the user visits the region, the bigger %the marker.} \label{fig:top_locations}
%\end{figure}

%
%
\section{Conclusions and Future Work}
\label{sec:ConclusionFutureWork}

%%%%%%%%%%%%%Conclusion%%%%%%%%%%
A historical record of the daily mobility pattern of the users hides an unexpectedly high degree of potential predictability despite the apparent randomness of human nature. Following this idea, we show that most people have a relatively regular schedule of moments when they travel from one location to another.

In this research, we present a new density-based clustering method to filter mobility data finding the most frequent places of a given individual and compare our method with two other proposals and show that this approach provides more informative results for this context. We also explore a new GSM dataset of diverse cities in Brazil showing the usefulness of the proposed clustering method to identify meaningful places over data with different granularity. We also introduce a Gaussian Mixture Model to find individuals' habits from the clustered data in a dynamic way.
%%%%%%%%%%%%%Conclusion%%%%%%%%%%

%%%%%%%%%%%%%FutureWork%%%%%%%%%%
For future work, we intend to propose a method to find the patterns of people visiting and leaving different places at different times in an order (weekly basis, daily basis) similar to sequential pattern mining methods. Also includes some map matching tasks including external information in order to find the semantic meaning of the individuals' movements. We also intend to apply the method in other datasets to verify its usefulness generalizing in other scenarios. Location prediction is also a field that is considered the results of this paper are strongly related to it.
%%%%%%%%%%%%%FutureWork%%%%%%%%%%
\section*{Acknowledgement}
This work is financed by National Funds through the Portuguese funding agency, FCT - Funda\c{c}\~ao para a Ci\^{e}ncia e a Tecnologia within project : UID/EEA/50014/2019

\bibliographystyle{splncs04}
\bibliography{biblio} 
\end{document}